\documentclass[twocolumn, 10pt]{article}

\usepackage{2013ConferencePHMSociety}
\usepackage{graphicx}
\usepackage{amsmath}

\begin{document}

\title{  An Offline Deep Reinforcement Learning  for Maintenance Decision-Making   }

\author{%
	Hamed Khorasgani
	\and 	Haiyan Wang
	\and 	Chetan  Gupta
	\And 	Ahmed  Farahat
}

\address{
	\affiliation{}{Hitachi Industrial AI Lab, Santa Clara, USA}{ 
		{\email{firstname.lastname@hal.hitachi.com}}
		} 
}

\maketitle
\pagestyle{fancy}
\thispagestyle{plain}

\phmLicenseFootnote{Hamed Khorasgani}

\begin{abstract}Several machine learning and deep learning frameworks have been proposed to solve remaining useful life estimation and failure prediction problems in recent years.  Having access to the remaining useful life estimation or  likelihood of failure in near future helps operators to assess the operating conditions and, therefore, provides  better opportunities  for sound repair and maintenance decisions. However, many operators believe remaining useful life estimation and failure prediction solutions are incomplete answers to the maintenance challenge. They  argue that knowing the likelihood of failure in the future is not enough to make maintenance decisions that minimize  costs and keep the operators safe. In this paper, we present a maintenance framework based on offline supervised deep reinforcement learning that instead of providing information such as likelihood of failure, suggests actions such as ``continuation of the operation" or ``the visitation of the repair shop" to the operators in order to maximize the overall profit. Using offline reinforcement learning makes it possible to learn the optimum maintenance policy from historical data without relying on expensive simulators. We demonstrate the application of our solution in a case study using the NASA C-MAPSS dataset.
 \end{abstract}

\section{Introduction}
\label{sec:electr-subm} 

Artificial intelligence (AI) has revolutionized  maintenance operations by providing more accurate predictive tools such as failure prediction and remaining useful life (RUL) estimation in recent years \cite{zheng2017long}. However,    industries  often  rely on human judgment to use  predictive tools and  determine maintenance decisions. 
\citeA{silver2016mastering}  have shown that deep reinforcement learning (RL) can achieve   a performance superior to human judgment in games such as chess and Go.  The main challenge of training deep RL for predictive maintenance decision-making is the  lack of reliable simulators in most industries \cite{khorasgani2020challenges}. Failure is often rare and complex. Therefore, it is challenging  to develop a simulator that can model these events. 

Recently, several researchers have shown that deep RL  can be formulated as a supervised problem. \citeA{schmidhuber2019reinforcement} introduced Upside Down RL (UDRL). Unlike traditional 
RL algorithms wherein an agent learns from  reward, UDRL learns a mapping from  desired  rewards and  observations  to  actions. The main  idea of UDRL is that supervised learning can learn a generalized  model  that can generate actions which can achieve requested rewards within requested  time. In UDRL,  learning includes two parallel algorithms: 1) A1 performs execution and exploration and 2) A2 performs  supervised learning. A1 generates actions   based on known maximum possible cumulative rewards using the current version of the model trained by A2. For the sake of exploration,  random actions are  taken   occasionally. The observations, actions and rewards are saved in a buffer. A2 uses historical data that is  extracted from the buffer  to train the model that maps the observation and maximum possible cumulative reward to optimal action.

\citeA{zha2021simplifying}  show that supervised learning can be competitive to state of the art RL  algorithms with
better  stability  and lower training  time.
\citeA{chen2021decision} introduced Decision Transformer which  expands UDRL to offline RL. Decision Transformer uses transformer architecture \cite{vaswani2017attention} to learn  a mapping from  historical observations, historical actions, and expected reward to proper actions using offline data. 
\citeA{janner2021reinforcement}  also show that a supervised approach based on the transformer architecture  is competitive to  other offline RL algorithms without the requirement of learning the system dynamic model or  even the value function. 
Relying purely on offline data makes offline RL  a perfect choice for maintenance decision-making wherein storing historical data to learn a policy is much more cost-effective   than developing a simulator that can model failure accurately. Using a supervised approach simplifies  the solution even more.  In this work, we use a supervised offline RL approach to learn optimal maintenance decision-making.  

The rest of the paper is as follows. Section \ref{sec:Approach} presents our approach. Section \ref{sec:Experimental} presents the experimental results and Section \ref{sec:Conclusion} concludes the paper.

\section{Maintenance Decision-Making As a Supervised RL Problem}
\label{sec:Approach}

We define the problem of maintenance decision-making as learning a  maintenance policy that generates optimal maintenance actions such as  ``continuation of the operation" or ``the visitation of the repair shop" based on the observations and the previous maintenance actions. The observations can come in different formats. For example, they may include time-series sensor data such as temperature, and  event data such as  the check engine light.  
A maintenance action is optimal if it leads to the maximum profit. The profit is the difference between the generated revenue and the  operating costs. The  operating costs include both maintenance  costs and  failure costs. Failure costs are often much higher than maintenance  costs because they occur without  planning. More importantly, in some cases, a failure can put the operator's safety in danger. In this paper, we  rely solely upon offline data to learn the maintenance policy. This makes our approach scalable and  practical as it does not require the development  and maintenance of a simulator. 

Consider the case wherein offline data includes the observations, $o_k$, actions, $a_k$ and reward $r_k$ at each given time $k$. A trajectory $i$  includes the set of observations over time, the set of actions taken over time,  and the set of rewards gained over time. For a given window with the length of  $T$, we can extract the $T$ steps' observation history of the equipment at each given time, $k$, as:
\begin{equation}
O^T_k= \{o_{k-T+1},..., o_k\},
\label{eq:ol}
\end{equation}
using the equipment trajectory. 
We can also extract $T$ steps' action history of the equipment at each given time, $k-1$, as:
\begin{equation}
A^T_{k-1}= \{a_{k-T},..., a_{k-1}\}.
\label{eq:a}
\end{equation}
Note that  unlike the observation history where we included $o_k$, here, we consider $a_{k-1}$ as the last action. This is because our goal is to use the  observation history and the action history to predict action  $a_k$. 

Consider horizon $H$. We can also calculate future cumulative rewards for each trajectory from time $k$ up to the horizon, $H$,  as $r^H_k=\sum_{i=k}^{k+H} r_i$ using  offline data. Calculating $r^H$ over the past $T$ time steps, we have:
\begin{equation}
R^H_k= \{r^H_{k-T+1},..., r^H_{k}\}.
\label{eq:r}
\end{equation}

Next, we use  a supervised learning method to learn a model which predicts action $a_k$ using  $O^T_k$, $A^T_{k-1}$, $R^H_k$.   
\begin{equation}
\hat{a}_k=model(O^T_k, A^T_{k-1}, R^H_k)
\label{eq:model}
\end{equation}
The idea here is to solve maintenance decision-making using an offline supervised  RL approach similar to Decision Transformer 
 \cite{chen2021decision} and Trajectory Transformer \cite{janner2021reinforcement}. In the training stage, we learn a mapping from  future expected  cumulative rewards to the actions at each state using offline data. In the application, we feed the model with high future expected cumulative rewards and we expect the model to generate actions to achieve such.  

\begin{figure}[t]
\centering
\includegraphics[scale=.34]{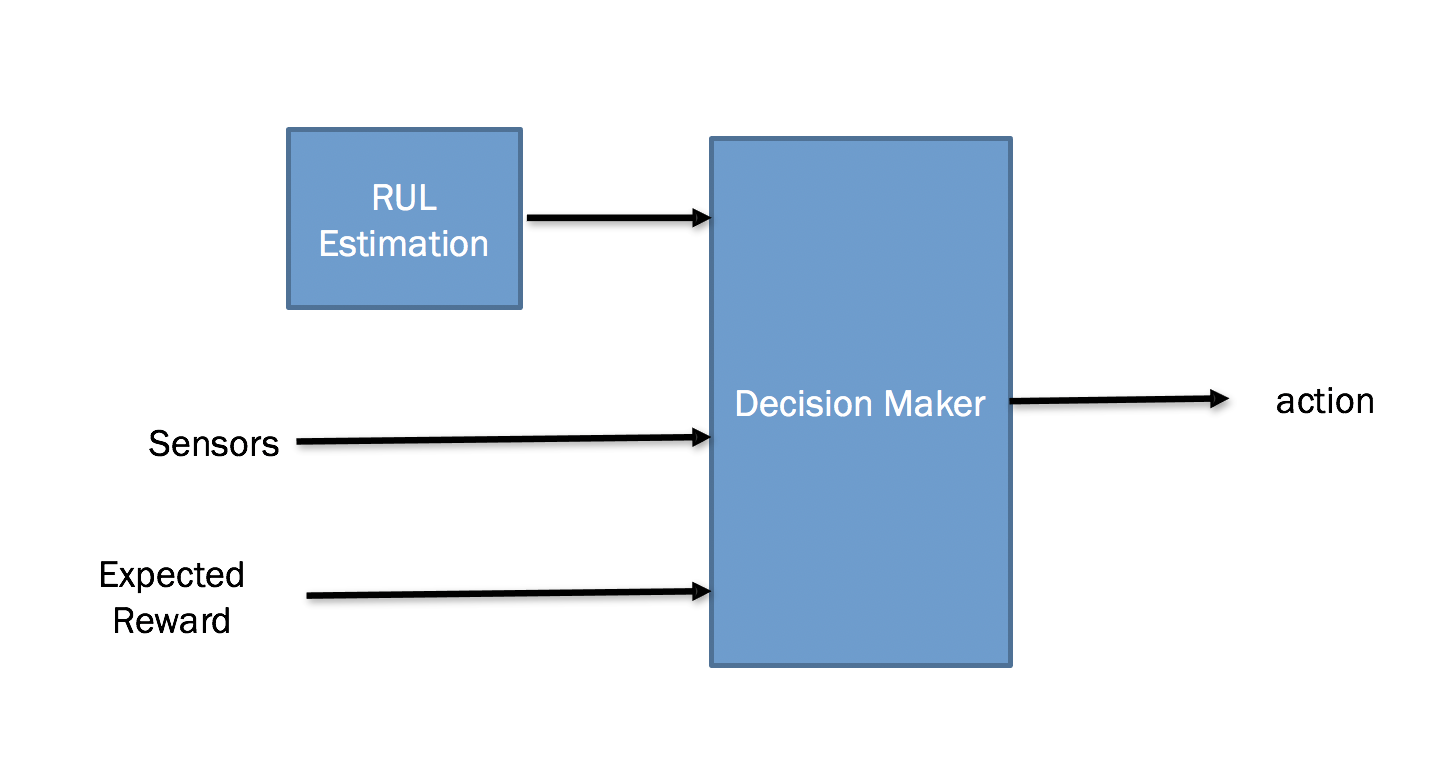}
\caption{Proposed architecture for smart maintenance decision-making.}
\label{fig:model}
\end{figure}

To design a reliable maintenance decision-making  model, we have to consider the following points: 
\begin{itemize}
\item \textit{RUL estimation and failure prediction models:} 
when enough data is available, our proposed maintenance decision-making framework can learn an optimal policy from the data  without requiring  a separate RUL estimation   or failure prediction unit. However, it is very common for  industries to  develop reliable RUL estimation models that can estimate the remaining useful life of   equipment. 
When these models are already available, their output can be used as an input to our decision-making model to improve accuracy and simplify the training process as it is shown in Figure \ref{fig:model}.  The formulation is:
 \begin{equation}
\hat{a}_k=model(O^T_k, A^T_{k-1}, R^H_k,RUL_k),
\label{eq:model2}
\end{equation}
 where $RUL_k$ presents the remaining useful life at time $k$. Note that during the training, we may have access to the actual RUL. However,   during the application process, we have to use the RUL estimation.

\item \textit{Reward function:}
 the goal of predictive maintenance is to minimize the operation costs and maximize the profit. Typically,  an equipment generates profit as it operates. So to maximize the profit, the industries aim to maximize equipment utilization. On the other hand, failures can be very expensive to fix or even dangerous for the operators. Therefore, operating  equipment to the point of failure is often not a viable option. The cost of maintenance and repair is another factor that we have to consider in maintenance decision-making. In addition to the extraneous labor and parts costs, unnecessary maintenance  reduces industries' profits by lowering  utilization time. Finally, the repair cost can depend upon the state of the equipment. For example, for some equipment, early fixes may be cheaper than late-stage repairs. Designing a reward function that can capture these complexities associated with costs and profits is crucial in designing an RL algorithm for maintenance decision-making. Fortunately, in many industries, the cost and profit of equipment operations are recorded in details and therefore can be used to design the reward function.

 \item \textit{Total expected reward:} as it is shown in equation (\ref{eq:model})  and (\ref{eq:model2}), the expected reward is an input to our model. During the training, we use the actual costs to compute the future rewards for each trajectory. During the application,   we must feed high expected future rewards so that the model generates good actions. Naturally, one may ask  ``is there a limitation on how high the  expected future rewards can be during the application?"  The answer to this question is  yes. Our experimental studies  in this paper  and previous experiments done by \cite{chen2021decision} show that after a certain point, increasing the expected future rewards has an opposite effect and degrades the performance. Our supervised offline RL algorithm uses historical data to learn the mapping from expected rewards to the actions at each state. Therefore, if during the application we feed the model with an expected future reward significantly outside of the training  dataset distribution, the model may generate unreliable  actions.

 \citeA{chen2021decision} show that the total expected rewards  
and the actual total returns are highly correlated as long as we feed the model with the total expected reward in the range which was observed in the offline data.  More interestingly, they show that 
it is even  possible to achieve  higher returns than the maximum episodic return 
in the dataset for certain specific tasks. This shows that offline supervised RL can   extrapolate accurately to some extent.
In this paper, we also observe  a high correlation between the total expected reward 
and the  observed total reward for the maintenance problem.  Moreover, our experiments  demonstrate  that selecting the  total expected reward    within     the vicinity of the maximum episodic return  available
in the offline dataset is a good practical approach to set this parameter  in the maintenance application.

\item \textit{Architecture:}
\citeA{chen2021decision}  and  \citeA{janner2021reinforcement}  used the transformer architecture \cite{vaswani2017attention} to learn the model that predicts the maintenance action (see  equations (\ref{eq:model}) and (\ref{eq:model2})). However, their approach can be implemented  using any  classifier  model when we have discrete action space. Similarly,   any regression model can be used to predict actions when we have continuous  action space. The  main parameters we have to consider in selecting the model  architecture are: 1) the amount of available data,   2) the complexity of the data, and 3) the available computational resources. In the experimental section, we will demonstrate  that a simple fully connected neural network architecture  can achieve acceptable performance for our dataset. 
 \end{itemize}

\section{Case Study}
\label{sec:Experimental} 

In this section, we use the C-MAPSS dataset to show the application of our proposed maintenance decision-making algorithm.  This dataset includes time-series sensor measurements of jet engine under  different operational conditions and fault modes. The dataset is generated by   the Prognostic Center of Excellence at NASA Ames using the C-MAPSS simulator \cite{saxena2008turbofan}.

We focus on equipment F002 in this experimental study. The training dataset for this equipment includes 260 trajectories. Each trajectory ends when the equipment fails. The test dataset includes 259 trajectories. Unlike the training dataset, the trajectories in the test dataset do not include failures. This was  mainly designed  to hide the failure points from the participants in the data challenge. To show the performance of our algorithm, when the system fails, we consider the first 250 trajectories in the original training dataset as our new  training dataset and the last 10 trajectories as our first test dataset. We consider the original test dataset as the second test dataset in this paper. 

\subsection{Removing the effect of operation modes}

Similar to  previous work \cite{wang2019remaining}, we first normalize sensor variables with respect to  the operation modes. The goal here is to remove the effect of different operating modes from the sensor data so that we can learn a maintenance decision-making model that works for all operating modes. Equations (\ref{eq:norm}) and (\ref{eq:norm2}) represent the normalization process. 
 \begin{equation}
\hat{s}_i =\text{model}_i(\text{operation modes})
\label{eq:norm}
\end{equation}
where $\text{model}_i$ represents the regression normalization model for sensor variable $s_i$ and $\hat{s}_i$ represents the estimation of $s_i$ made by its normalization model  using three operating modes as the inputs. Note that the equipment has  21 sensor variables and we train a regression normalization model for each sensor.  We normalize each sensor variable $s_i$ using  its estimation $\hat{s}_i$ as follows:
 \begin{equation}
s_{in}=\frac{s_i}{\hat{s}_i}
\label{eq:norm2}
\end{equation}
where $s_{in}$ represents normalized $s_i$. 

\begin{figure}[t]
\centering
\includegraphics[scale=.34]{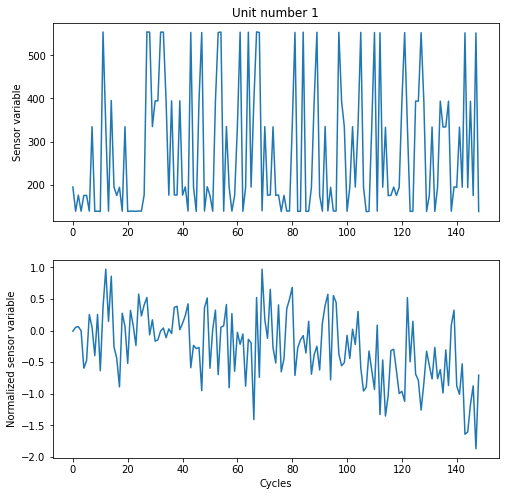}
\caption{An original sensor value, $s_i$, and its normalized value, $s_{in}$, show that normalization amplifies  the effect of degradation on sensor variables by removing  the effect of operation modes.}
\label{fig:sensor}
\end{figure}
Figure \ref{fig:sensor} shows a sensor variable  before and after normalization. We can see   the degradation process in the sensor variables when the effect of operation mode is removed.

\subsection{Generating a  dataset for training   decision-making model}

Unfortunately, the C-MAPSS dataset does not include any maintenance action. In the original training dataset, during the simulation, all of the equipment has been utilized to the point of failure. In the test dataset, the simulation has stopped before the failure without taking any  maintenance action. 
 To train  our models (see equations (\ref{eq:model}) and (\ref{eq:model2}) ), we must have access to historical actions. We modify the dataset by assuming two kinds of actions: 1) continue operation; 2) repair the equipment. The original dataset only includes continual operation (no actions). We consider random repairs in the dataset. We assume that after each repair, the equipment will return to its initial condition.

\subsection{Costs and rewards}

In this paper, we consider a  simple cost structure.   In our setting, each failure costs $250 + U(-50,50)$; each repair costs $25+  U(-5,5)$; and  equipment generates $1 +  U(-0.2,0.2) $ profit per operating cycle, where   $U(a,b)$ represents uniform distribution between $a$  and $b$. 
 Note that the ratio between the cost of failure and repair defines how conservative the network will be in taking proactive maintenance action. In an extreme case where the repair cost is equal to or more than the cost of failure, the optimal solution would be to operate the equipment to the point of failure. In another extreme case where  repair costs nothing, the optimal solution would be to repair the equipment after each cycle. 
 In practice, these numbers should be derived from facts on the ground. It is also possible that the costs change based on operation modes and the state of the equipment. For example, for a truck, it may be cheaper and faster to do repair and maintenance in a large city compared to a remote small town where  expert operators and equipment parts may be scarce.

\subsection{RUL estimation}
\begin{figure*}[t]
\centering
\includegraphics[scale=.55]{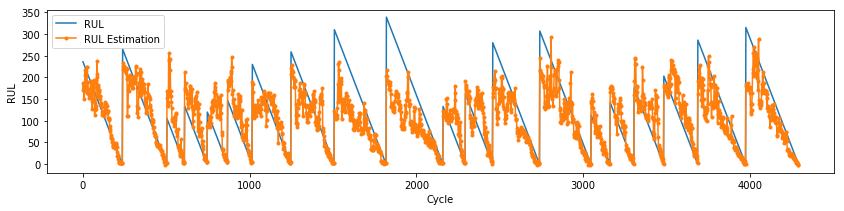}
\caption{RUL vs RUL estimation for our first test dataset (unit number 251 to 260). $R^2 = 0.53$.}
\label{fig:RUL}
\end{figure*}
As we mentioned earlier, we divided the equipment F002 dataset to train and test. We considered unit numbers  one to 250 as our new training dataset, and  unit numbers 251 to 260 as our first  test dataset. We considered the original equipment test dataset as our second test dataset. 
We then used the training dataset to learn a normalization model for each sensor. We use the long short-term memory (LSTM) architecture proposed by  \cite{zheng2017long} for the RUL estimation
 using normalized sensors. 

\begin{figure*}[t]
\centering
\includegraphics[scale=.55]{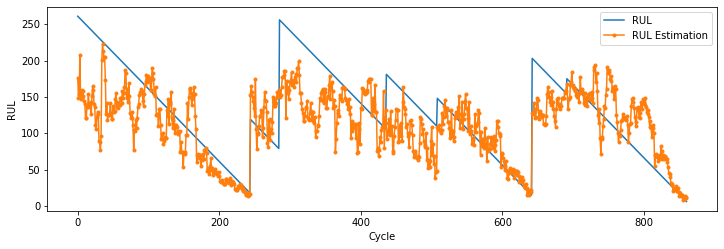}
\caption{RUL vs RUL estimation for second test dataset using LSTM neural networks. $R^2 = 0.42$}
\label{fig:RUL_LSTM}
\end{figure*}

  Figure \ref{fig:RUL} shows the RUL estimation for our first test dataset. Figure \ref{fig:RUL_LSTM} shows the RUL prediction  for the first 6 trajectories in our second test dataset. We  see that the RUL estimation becomes more accurate as it gets closer to the equipment's end of life. 
Our first test  dataset (trajectories 251 to 260 of the original training dataset) has a more similar distribution to our training dataset (trajectories one to 250 of the original training dataset) and therefore our model predicts RUL more accurately for this dataset compared to the second test dataset ($R^2 = 0.53$ vs $R^2 = 0.42$). 

\subsection{Offline RL}

In this subsection, we train a   multilayer perceptron (MLP)  neural network  to predict one of these two actions: 1) continue operation or 2) repair the equipment, based on the expected future rewards. Our network has one hidden layer with 100 neurons. We use the rectified linear activation function (ReLU),  we apply the Adam optimization and set the  learning rate  equal to $0.001$. 
We train two models: in the first model, we only use sensor data and the  expected rewards to predict the action. In the second model, we add RUL as an additional feature. For the model with RUL, we use actual RUL during the training and the estimation of RUL during the test time.

\subsection{Application}

\begin{figure}[t]
\centering
\includegraphics[scale=.28]{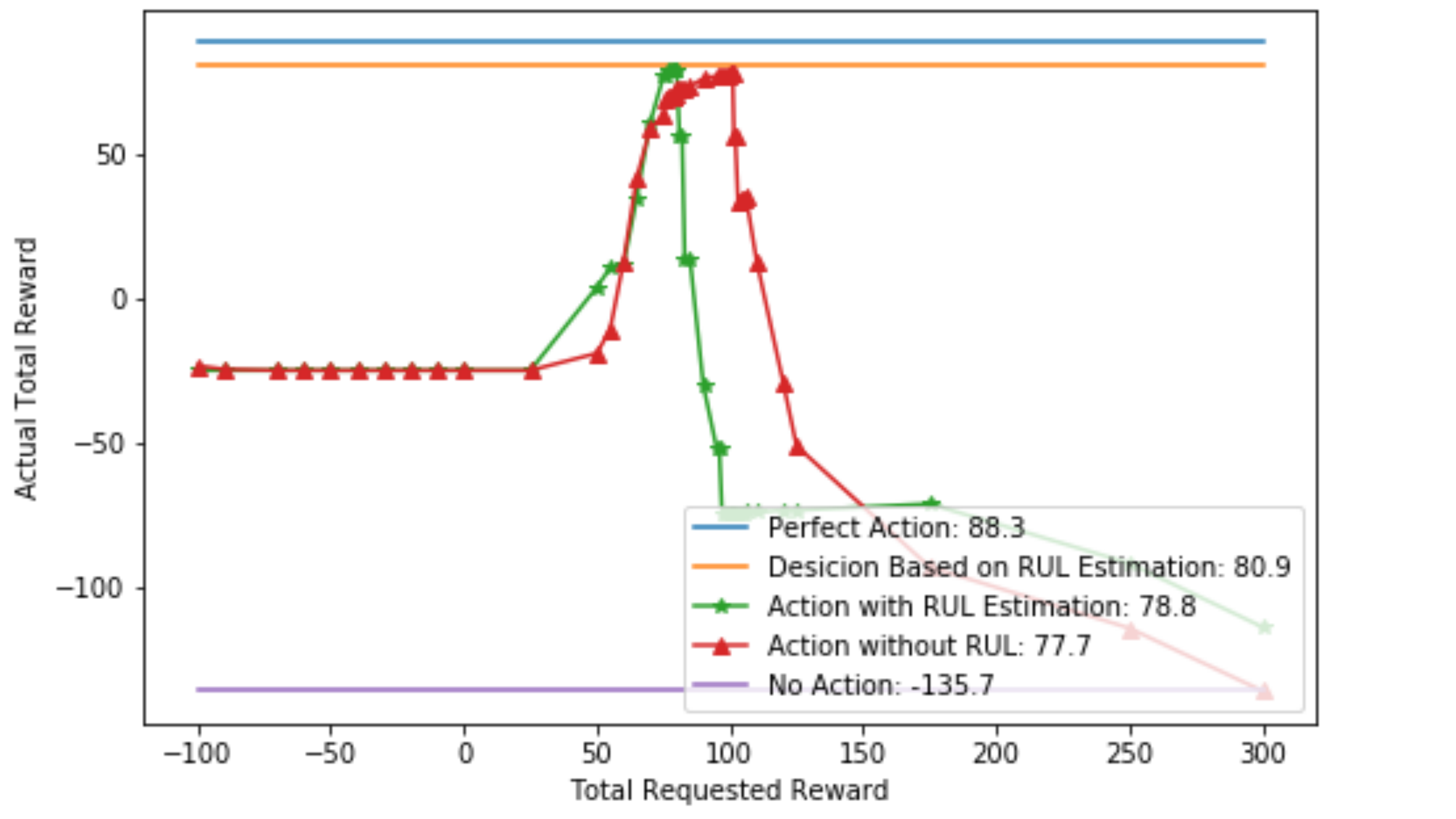}
\caption{Average cumulative rewards as a function of expected reward for the first test dataset. The maximum average cumulative rewards for the  model with RUL achieves at the total expected reward =  79, and the maximum average cumulative rewards for the model without RUL achieves at the total expected reward =  100.5.}
\label{fig:results1}
\end{figure}
\begin{figure}[t]
\centering
\includegraphics[scale=.28]{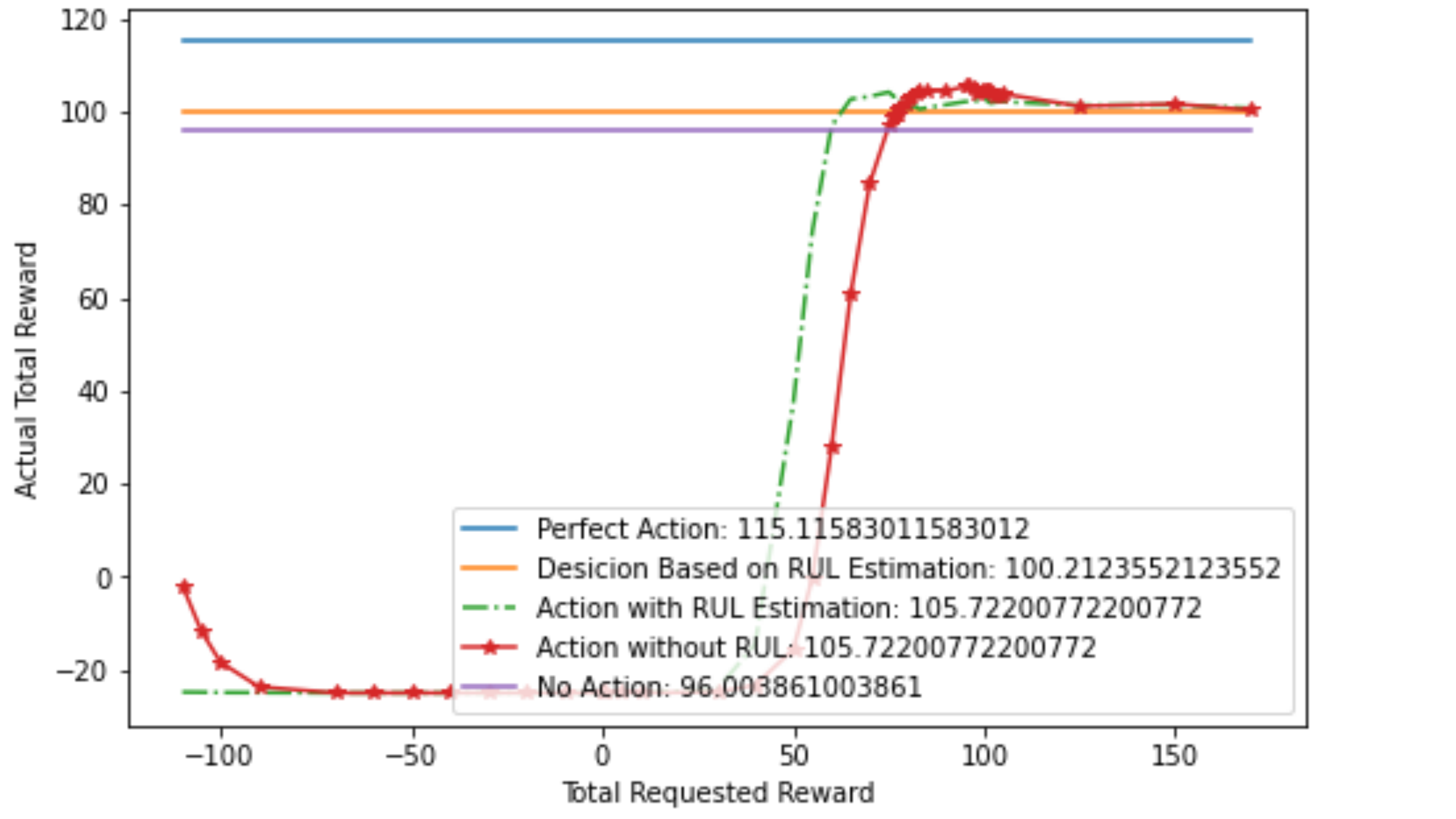}
\caption{Average cumulative rewards as a function of expected reward for the second test dataset.The maximum average cumulative rewards for the  model with RUL achieves at the total expected reward =  75, and the maximum average cumulative rewards for the model without RUL achieves at the total expected reward =  96.}
\label{fig:TestResults}
\end{figure}

We consider each experiment in the test dataset and episode. An episode  ends when 1) its trajectory ends, 2) the equipment  fails, or 3) the equipment goes for  repair. Each cycle of operation brings 1 unit of profit, and each repair costs 25 units. An early repair  lowers the profits by ending the episode. A failure costs 250 units. We assume that  repair decisions should be made at least 10 cycles before  failure occurs. This is equal to the time that the equipment requires to get to the repair shop or to stop  operation safely.   Figure \ref{fig:results1} shows that no action would cost -135.7 on average (see the purple line) for our first  test dataset. This is because the first test dataset includes failures and each episode without action would end with a failure. 
Obviously, allowing the equipment to operate  to the point of failure    is not an acceptable solution for any business. 
Figure \ref{fig:TestResults} shows that no action would earn  96 units of profit  on average in the second test dataset (see the purple line). This is because in the second dataset, the episode often ends before reaching its  failure point. 

The optimal solution is to take the equipment for repair 
 10 cycles before reaching  failure. In this case, we maximize  profit as we utilize the equipment to its full potential while  avoiding the high cost of failure. The blue line in Figure \ref{fig:results1} shows this scenario. We achieved 88.3 average cumulative reward using this approach for the first test dataset.   Of course, this is not a practical solution because it requires the knowledge of exact RUL. An alternative option would be to use  the RUL estimation instead  of the actual RUL. The orange line in Figure \ref{fig:results1} shows this scenario for the first test dataset. We achieved 82.7 average cumulative reward using this approach.

 Note that this approach is not fully data-driven as it requires the decision-making  logic of taking the equipment for repair 10 cycles before  failure. This logic is trivial for this simple example. However, when we have more complicated cases, it may not be that simple to come up with an optimal solution. For example, the repair cost may change based on the system  conditions. This may make our simple logic suboptimal as the repair 10 cycles  before  failure can be more expensive than earlier repairs.  It is also possible that we  have more than one part to repair. For these cases, offline RL presents a better alternative as it learns the optimal policy from data.

 We can see in Figure \ref{fig:TestResults} that the difference between the perfect action and the action based on RUL estimation is wider in the second dataset compared to the first test dataset. This is because the RUL estimation in the second dataset is  less accurate and thus makes decisions based on RUL estimation less reliable.

 Figure \ref{fig:results1} and Figure \ref{fig:TestResults} show the offline RL performance for two scenarios: 1) in the first one, we used real RUL during the training, and RUL estimation during the test as an input to our offline RL model, 2) in the second case, we trained the offline RL purely relying on sensor data.  The results show that the model achieves similar performance with or without  RUL estimation. 
  We believe that this is because in this dataset, the normalized sensors capture the degradation process  and having access to RUL estimation does not provide significant additional information for the model. 

Note that the  offline RL outperforms decisions  based on RUL estimation in the second test dataset (see Figure \ref{fig:TestResults}). Figure \ref{fig:TestResults} shows that the performance of the model with the RUL estimation peaked  at total requested rewards equal to 75 and the performance of the model without RUL estimation peaked at total requested rewards equal to 96. However, both models perform  better than the  decision based on RUL estimation for a fairly wide range of total requested rewards. 
This presents a huge potential for offline RL in maintenance decision-making, especially for real-life problems when estimating RUL is not a trivial task.   



\section{Conclusion}
\label{sec:Conclusion}
In this paper, we presented a framework to use offline reinforcement learning for maintenance decision-making. The results show  that offline RL can provide  decision-making  competitive  to using RUL estimation. Moreover, offline RL can generate acceptable solutions even  without requiring learning the RUL estimation model. Our approach provides a framework that can be applied to more complicated maintenance decision-making challenges.

\begin{singlespace} 
\bibliographystyle{apacite}
\bibliography{ijphm}
\end{singlespace} 




\end{document}